\documentclass{article}

\usepackage{amsmath}

\usepackage{arxiv}

\usepackage[utf8]{inputenc} 
\usepackage[T1]{fontenc}    
\usepackage[hidelinks]{hyperref}       
\usepackage{url}            
\usepackage{booktabs}       
\usepackage{amsfonts}       
\usepackage{nicefrac}       
\usepackage{microtype}      
\usepackage{float} 
\usepackage{lipsum}		
\usepackage{graphicx}
\usepackage[numbers]{natbib}
\usepackage{doi}
\usepackage{subcaption}

\title{A Comparative Study of light-weight Language Models for PII Masking and their deployment for real conversational texts
}

\author{
    \href{https://orcid.org/0009-0004-9526-9842}{Prabigya Acharya} \\
    IOE, Pulchowk Campus\\
    \texttt{076bct043.prabigya@pcampus.edu.np} 
    \And
    \href{https://orcid.org/0009-0003-2937-6837}{Liza Shrestha} \\
    IOE, Pulchowk Campus \\
    \texttt{076bct032.liza@pcampus.edu.np} 
}

\renewcommand{\shorttitle}{\textit{Discrypt}}

\hypersetup{
pdftitle={Discrypt},
pdfsubject={q-bio.NC, q-bio.QM},
pdfauthor={Prabigya Acharya, Liza Shrestha},
pdfkeywords={Personally Identifiable Information (PII), PII Masking, T5,
Mistral, Privacy-Preserving NLP},
}

\begin{document}
\maketitle

\begin{abstract}
Automated masking of Personally Identifiable Information (PII) is critical for privacy-preserving conversational systems. While current frontier large language models demonstrate strong PII masking capabilities, concerns about data handling and computational costs motivate exploration of whether lightweight models can achieve comparable performance. We compare encoder–decoder and decoder-only architectures by fine-tuning T5-small and Mistral-Instruct-v0.3 on English datasets constructed from the AI4Privacy benchmark. We create different dataset variants to study label standardization and PII representation, covering 24 standardized PII categories and higher-granularity settings. Evaluation using entity-level and character-level metrics, type accuracy, and exact match shows that both lightweight models achieve performance comparable to frontier LLMs for PII masking tasks. Label normalization consistently improves performance across architectures. Mistral achieves higher F1 and recall with greater robustness across PII types but incurs significantly higher generation latency. T5, while less robust in conversational text, offers more controllable structured outputs and lower inference cost, motivating its use in a real-time Discord bot for real-world PII redaction. Evaluation on live messages reveals performance degradation under informal inputs. These results clarify trade-offs between accuracy, robustness, and computational efficiency, demonstrating that lightweight models can provide effective PII masking while addressing data handling concerns associated with frontier LLMs.
\end{abstract}
\keywords{PII Masking, Large Language Models, Fine-tuning, Data Privacy, ROUGE Evaluation}

\section{Introduction}
Large language models (LLMs) have revolutionized natural language processing tasks like text generation, summarization, and question answering across healthcare, finance, and legal domains. However, user prompts often embed personally identifiable information (PII)\citep{singh2025, karavdic2025}. PII refers to any data that may be used to directly or indirectly identify an individual, such as names, addresses, social security numbers, phone numbers, and financial and health records. As data handled increases, accurate and scalable redaction of sensitive information is non-negotiable specially as mishandling personal data can have serious legal, ethical, and financial consequences \citep{leon2025}.  For example, HIPAA prohibits sharing electronic health records (EHRs) for research unless they are properly de-identified \citep{hipaa2025}.

Hence, PII masking is required to protect user privacy, ensure regulatory compliance, and prevent unintended data leakage during model training and inference. Prior studies have shown that language models can memorize and reproduce sensitive information present in their training data, posing serious privacy risks when deployed in real world applications \cite{carlini2021extracting, brown2020language}. In interactive settings, LLMs may inadvertently expose personal details through logs, prompts, or generated outputs, particularly when handling conversational, medical, or customer service data \cite{leon2025, karavdic2025}. As LLMs are increasingly integrated into high-stakes decision making pipelines, robust and automated PII masking becomes a critical safeguard for ethical, legal, and trustworthy AI deployment.

Traditional PII redaction methods, including regex patterns and statistical matching, offer speed but falter on contextual nuances essential for accurate masking across varied texts. These rigid tools struggle with format variations like differing phone number conventions between regions limiting scalability for global, multilingual applications. Commercial APIs from AWS Comprehend \citep{aws_comprehend_2025}, Microsoft Presidio \citep{microsoft_presidio}, and Google Cloud\citep{google_cloud_dlp_2025} have boosted accuracy with deep learning, yet remain black-box solutions requiring data transmission to external providers. This poses compliance risks under GDPR or HIPAA for sensitive sectors. Open-source LLMs like Mistral and T5 address these issues through inherent contextual awareness, enabling fine-tuned redaction that generalizes broadly while supporting self-hosted deployment for full control and transparency.

This study addresses the following research questions:
\begin{enumerate}
    \item How do encoder-decoder models (T5) compare with decoder-only LLMs (Mistral) in PII masking accuracy?
    \item How robust are these models across different PII types in conversational text?
    \item What trade-offs exist between model performance, output controllability, and computational cost?
\end{enumerate}
To this end, we make the following contributions:
\begin{itemize}
    \item We present a systematic evaluation of encoder-decoder and decoder-only architectures for PII masking under controlled dataset variants. We fine-tune a T5 and Mistral models on 3 datasets, with varying level of control in input data and target labels.
    
    \item We analyze the impact of dataset curation strategies using strict entity-level and character-level metrics. Our evaluation reveals how canonical normalization, regex-based replacement of unmasked PII, and removal of irrelevant PII candidates differentially affect robustness, boundary detection, and type accuracy.
    
    \item We evaluate model performance on real world data using a discord bot to read and censor PII-containing messages. 
\end{itemize}

\subsection{Related Works}
\subsubsection{PII masking}
Personally Identifiable Information (PII) is any information about an individual maintained by an agency, including any information that can be used to distinguish or trace an individual's identity, such as name, social security number, date and place of birth, mother's maiden name, or biometric records; and any other information that is linked or linkable to an individual, such as medical, educational, financial, and employment information \citep{nist800122_2010}. PII Masking is the systematic process of identifying these sensitive elements and applying transformations to hide the original values while attempting to preserve the data’s utility for downstream applications \citep{deshmukh2023lifepiipii}. To achieve PII masking, several methods have evolved, transitioning from static pattern matching to context-aware generative models.

\begin{enumerate}
\renewcommand{\labelenumi}{\Roman{enumi}.}

    \item \textbf{Regex Based Methods} \\
    Regular expressions(Regex) provide deterministic pattern matching for structured PII. Regex has been employed for general Named Entity Recognition tasks like automated resume parsing and high-risk PII detection. However, this approach suffers from high false positives and performs poorly on unstructured formats~\citep{rajgarhia2025_hybrid_pii}.       

    \item \textbf{Named Entity Recognition (NER)} \\
    Named entities are real-world objects represented by proper names or identities of interest. Named Entity Recognition (NER) and classification is essentially a sequence labeling task, where, given a sequence of tokens, a system aims to assign appropriate labels (NE classes) to each token~\citep{ehrmann2021ner_survey}. However, traditional NER approaches face several limitations in accurately identifying personally identifiable information (PII) due to the sensitivity and complexity of the content. Challenges include ambiguity, polysemy, context-dependence or domain-dependence, and variations in phrasing such as nicknames, aliases, informal expressions, alternative representations, emerging terms, evolving naming conventions, syntax variations, typos, and misspellings~\citep{singh2025_unmasking_pii}. Natural language processing platforms like OpenNLP and spaCy provide NER toolkits for identifying PIIs. Among the two, spaCy, a Python-based industrial-strength NLP library, outperforms OpenNLP on F-measure and recall for IME report de-identification ~\citep{pearson2021_ner_medical}.

    \item \textbf{Transformer-based Approaches}\\
    The transformer architecture, replaces recurrent layers with stacked self-attention and point-wise feed-forward networks, enabling parallelization and capturing long-range dependencies without sequential processing \citep{vaswani2017_attention} .
    \begin{itemize}
        \item \textbf{Encoder-Only Models}  Encoder-only architectures, exemplified by BERT \cite{devlin2018bert} and its successors like RoBERTa \citep{liu2019robertarobustlyoptimizedbert} and DeBERTa \citep{he2021debertadecodingenhancedbertdisentangled}, utilize bidirectional self-attention to generate context-aware representations for every token along with a few modifications. In the context of PII, these models treat masking as a sequence labeling or token classification task. Within this paradigm, DeBERTa v3 has emerged as a high-performance baseline for identifying sensitive entities due to its disentangled attention mechanism. This architectural innovation represents word content and relative positions using separate vectors, allowing the model to more effectively capture the structural nuances of language such as the relationship between a name and its surrounding context that are critical for accurate entity detection \citep{he2021debertadecodingenhancedbertdisentangled}.
        \item \textbf{Encoder-Decoder Models} The Encoder-Decoder (or sequence-to-sequence) architecture, pioneered by models like T5 \citep{raffel2023exploringlimitstransferlearning} and BART \citep{lewis2019bartdenoisingsequencetosequencepretraining}, diverges from the discriminative approach by treating PII masking as a generative task. In this framework, the encoder processes the input sequence to create a latent representation, which the decoder then uses to generate an entirely new, transformed string \citep{raffel2023exploringlimitstransferlearning}.
        
        \item \textbf{Decoder-Only Models} The rise of large decoder-only models, such as Mistral \citep{jiang2023mistral7b} and Llama \citep{touvron2023llamaopenefficientfoundation}, has introduced a new paradigm for "zero-shot" and "few-shot" PII redaction. These autoregressive models utilize vast pre-training to understand nuanced, domain-specific contexts that traditional NER might miss. The PRvL (PII Redaction via LLMs)\citep{leon2025} paper provides a comprehensive benchmark for this class of models, establishing an open-source suite of fine-tuned LLMs specifically for general-purpose PII redaction. 
    \end{itemize}
\end{enumerate}

\subsubsection{Datasets}
The availability of high-quality annotated datasets is crucial for developing and evaluating PII masking systems. Early work in this area primarily relied on synthetic or rule-based datasets derived from Named Entity Recognition (NER) corpora, such as CoNLL-2003, where a limited subset of entities (e.g., names and locations) could be interpreted as PII \citep{tjong2003conll}. However, these datasets lack comprehensive coverage of real-world PII types and do not reflect the diversity and contextual sensitivity required for practical privacy-preserving applications.

To address these limitations, more recent efforts have focused on creating datasets explicitly designed for PII detection and redaction. The ai4privacy dataset provides large-scale, automatically generated and human-in-the-loop PII annotations across diverse document styles, covering a wide range of PII categories including names, contact information, identifiers, and financial data \citep{ai4privacy_pii_masking_200k}.

\subsubsection{Metrics}
There are various metrics that can be used for the evaluation of PII masking, each having its own specific significance and interpretations
\begin{enumerate} \renewcommand{\labelenumi}{\Roman{enumi}.}

\item \textbf{Strict vs. Relaxed Evaluation} \\
Following the PRvL framework \citep{leon2025}, we evaluate detection performance using two criteria:
\begin{itemize}
    \item \textbf{Strict Evaluation:} A predicted PII entity is considered correct only if both the entity type and the exact character-level boundaries match the ground truth.
    \item \textbf{Relaxed (Partial) Evaluation:} A prediction is considered correct if there is any overlap between the predicted boundaries and the ground truth, provided the entity type is correct \citep{leon2025}.
\end{itemize}

\item \textbf{Classification Metrics: Precision, Recall, and F1-Score} \\
These are the standard metrics for the sequence labeling aspect of PII masking. 
\begin{itemize}
    \item \textbf{Precision:} Measures the proportion of identified PII that were actual PII (minimizing false positives).
\begin{equation}
    \text{Precision} = \frac{TP}{TP + FP}
\end{equation}
    \item \textbf{Recall (Sensitivity):} Measures the proportion of actual PII that were correctly identified. In privacy contexts, Recall is often prioritized to ensure no sensitive data "leaks" through the mask.
\begin{equation}
    \text{Recall} = \frac{TP}{TP + FN}
\end{equation}

\item \textbf{Accuracy:} Measures the proportion of correct predictions among all predictions made by the model.
\begin{equation}
    \text{Accuracy} = \frac{TP + TN}{TP + TN + FP + FN}
\end{equation}
\end{itemize}
Here, we define:
\begin{itemize}
    \item \textbf{TP}: True Positive
    \item \textbf{TN}: True Negative
    \item \textbf{FP}: False Positive
    \item \textbf{FN}: False Negative
\end{itemize}
\item \textbf{Utility and Similarity Metrics: ROUGE and BLEU} \\
To measure how much the masking process alters the original text's meaning and syntax, we utilize:
\begin{itemize}
    \item \textbf{BLEU (Bilingual Evaluation Understudy):} Originally for translation, it measures n-gram overlap between the original and masked text to ensure grammatical structure remains intact \citep{papineni2002bleu}.
   \item \textbf{ROUGE (Recall-Oriented Understudy for Gisting Evaluation):}
ROUGE measures the n-gram overlap between the model-generated output and the reference text to assess content preservation. We report three variants:
\begin{itemize}
    \item \textbf{ROUGE-1:} Measures unigram (single-word) overlap.
    \item \textbf{ROUGE-2:} Measures bigram (two-word sequence) overlap.
    \item \textbf{ROUGE-L:} Measures the longest common subsequence (LCS) between the hypothesis and reference, capturing structural similarity.
\end{itemize}

Each variant is reported using the F1 score:
\begin{equation}
    \text{ROUGE-F1} = \frac{2 \times \text{Precision} \times \text{Recall}}{\text{Precision} + \text{Recall}}
\end{equation}

where precision and recall are computed based on overlapping n-grams (or LCS for ROUGE-L) between the hypothesis and reference \citep{leon2025}.

\end{itemize}

\item \textbf{Semantic Privacy Metric: SPriV} \\
The \textbf{SPriV (Semantic Privacy)} metric, introduced in the PRvL paper, is a specialized metric for LLM-based redaction. It evaluates the semantic similarity between the redacted output and a "gold standard" redaction using embedding-based similarity (e.g., Cosine similarity of sentence embeddings). This ensures that even if the words differ, the \textit{intent} and \textit{privacy level} of the output remain consistent with requirements \citep{leon2025}.
\end{enumerate}

\section{Method}
We define the PII masking task same as in PRvL \cite{leon2025}. Given an input sentence, denoted by \begin{math} x = [x_{1}, x_{2},...,x_{n} ] \end{math}, where each token \begin{math}
x_i    \end{math} is associated with a label \begin{math} l_i \in \left\{0, 1  \right\} \end{math} indicating if it is a span of PII. Our goal is to generate the output sentence \begin{math} y = [y_{1}, y_{2},...,y_{n} ] \end{math} such that any corresponding PII in $x$ is replaced by the appropriate PII mask in $y$.\\

Furthermore, we denote a predefined set of PII entity types as \begin{math} T = [t_{1}, t_{2},...,t_{n} ]\end{math}, such as \texttt{PERSON}, \texttt{EMAIL}, or \texttt{LOCATION}. For each $x_i$ where $l_i = 1$, the corresponding output $y_i$ is drawn from a standard list of mask tokens $T$. Thus, the transformation for a PII token $x_i$ of type $t_j$ can be expressed as: \begin{equation}
y_i = 
\begin{cases} 
    [t_j] & \text{if } l_i = 1 \\
    x_i & \text{if } l_i = 0 
\end{cases}
\end{equation}

where $[t_j]$ is a mask token selected from the token set ${T}$.

Building on the formal definition of the PII masking task, we employ two state-of-the-art transformer architectures: T5 (Text-to-Text Transfer Transformer) and Mistral.

\subsection{Dataset}
The primary data source for this study is the ai4privacy/pii-masking-200k dataset. For the scope of this research, we filtered the corpus to include only English language instances.

\subsubsection{Dataset Audit}
To understand the underlying distribution and quality of the data, we conducted an audit of a randomly sampled subset of 1,000 instances. This audit revealed a high density of privacy-sensitive information, with an average of 21.45 masked entities per sample across more than 20 distinct PII categories, indicating a dense and diverse privacy-sensitive dataset.\\ \\
Expanding this audit to the full training set (29,908 samples) confirmed the dataset's complexity, yielding 385,284 mask tags at an average of 12.88 masks per sample. However, this preliminary analysis identified 225 unique tags, many of which exhibited label redundancy and heuristic leakage.\\ \\
1150 failures were detected, with the highest being in the \texttt{[EMAIL]} category, which were detected by creating standard regex mapping for common PII like email, ip-address, phone number, and so on. Furthermore, we observed multiple label variants for semantically similar entities (\texttt{[TEL]} vs \texttt{[TEL\_B]}), reflecting inconsistencies in annotation, while some tags like \texttt{[XXXX]}, \texttt{[EOF]} made no sense at all or just repeated the tags in the input, we consider these as errors in the annotations.

\subsubsection{Heristic-Based Baseline Evaluation}
To establish a definitive baseline and identify systematic errors in the original annotations, we implemented two heuristic-based detection methods: a custom Regular Expression (RegEx) suite and Named Entity Recognition (NER) using SpaCy’s en\_core\_web\_sm model.

The RegEx approach targeted high-precision patterns for entities such as emails, IP addresses, and credit card numbers. By mapping regex outputs against the target text, we identified 1,150 total failures where PII remained unmasked, including 558 instances of emails and 198 phone numbers. As shown in \ref{table:baseline_comparison}, the heuristic methods struggled significantly with recall and F1 scores.

\begin{table}[h]
\centering
\caption{Performance Comparison of Heuristic Baseline Methods}
\label{table:baseline_comparison}
\begin{tabular}{@{}lcc@{}}
\toprule
\textbf{Metric Category} & \textbf{RegEx Baseline} & \textbf{SpaCy (NER) Baseline} \\ \midrule
\textbf{Character-Level} &  &  \\
\quad Precision & 0.2360 & 0.1659 \\
\quad Recall & 0.1741 & 0.1227 \\
\quad F1 Score & 0.1911 & 0.1310 \\
\quad Exact Match Rate & 0.1241 & 0.0503 \\ \midrule
\textbf{Span-Based} &  &  \\
\quad Average IoU & 0.2106 & 0.1216 \\ \midrule
\textbf{Entity-Level} &  &  \\
\quad Precision & 0.2333 & 0.1020 \\
\quad Recall & 0.1257 & 0.0595 \\
\quad F1 Score & 0.1511 & 0.0675 \\
\quad Type Accuracy & 0.1683 & 0.0576 \\ \bottomrule
\end{tabular}
\end{table}

\subsubsection{Dataset Normalization}
\begin{enumerate}
\item{\textbf{Cannonical Mapping}}\\
Building upon the definition of the PII masking task, we introduce a normalization layer to ensure label consistency and data quality. We have defined $T$ as the set of PII entity types. Let $L_{raw} = T$ be the set of raw labels sampled from the dataset.
We observe that many labels $l \in L$ are semantically redundant.\\

We define a canonical mapping function $f_{norm}: L_{raw} \to M$, which transforms a diverse set of label variants into a unified canonical set $M$, where $|M| = 24$.\\

For any label $l_i$ associated with token $x_i$, the normalized label $l'_i$ is given by:\\ $$l'_i = f_{norm}(l_i)$$\\
For example, the subset of labels representing person entities \\
$$\{[\text{GIVEN\_NAME\_1}], [\text{FIRST\_NAME}], [\text{LAST\_NAME}]\} \subset L_{raw}$$ 
is mapped such that:$$f_{norm}(l) = [\text{PERSON\_NAME}] \quad \forall l \in \{[\text{GIVEN\_NAME\_1}], [\text{FIRST\_NAME}], [\text{LAST\_NAME}]\}$$\\

The list of supported pii lablels after mapping is in Table \ref{tab:pii_labels}

\item{\textbf{Heuristic-Based Correction}}:\\
To address the heuristic leakage identified in Section 2.1.1, we apply a correction function $h(x, y)$ based on a set of Regular Expressions $R$. If a pattern $r \in R$ identifies a PII span in the input $x$ that is not masked in the target $y$, the target is updated:$$y'_i = \begin{cases} [m_j] & \text{if } \exists r \in R \text{ s.t. } r(x_i) = \text{true} \\ y_i & \text{otherwise} \end{cases}$$This ensures that leaked entities (e.g., specific email formats or IP addresses) are properly represented by their canonical mask tokens in the training data.

\item{\textbf{Filtering}}:\\
Finally, we impose a constraint on the label space to prioritize the most relevant privacy categories. Let $T_{ranked} = \{t_1, t_2, \dots, t_{225}\}$ be the set of entity types ordered by their frequency in the audited dataset. We define a restricted taxonomy $T^* = \{t_1, \dots, t_{24}\}$, consisting of the top 24 most frequent entity types.The final dataset $D_{final}$ is constructed by filtering the samples such that for any sample $S = (x, y)$, the sentence is retained only if:$$\forall y_i \in y, \quad y_i \in M_{T^*} \cup \{x_i\}$$where $M_{T^*}$ is the set of mask tokens corresponding to the truncated taxonomy $T^*$. Sentences containing tags outside this set were removed unless they could be explicitly mapped to a canonical tag within $T^*$.
\end{enumerate}

\subsection{Finetuning T5}
To adapt the T5 model for the PII masking task, we employ a supervised fine-tuning approach using a modified and filtered version of the AI4Privacy dataset. We select the T5-small architecture to balance computational efficiency and sequence generation quality, making it suitable for low-latency conversational environments.

Consistent with the text-to-text framework proposed by \citet{raffel2023exploringlimitstransferlearning}, each input instance is reformulated as a transformation task through the use of a task-specific prompt. This prompt is prepended to the input text, ensuring that the model interprets the task as sequence-to-sequence generation rather than standard language modeling. An example of the formatted input is shown below:
\begin{verbatim}
mask pii: <input_text>
\end{verbatim}

The model is fine-tuned for three epochs using the AdamW optimizer with a weight decay of $0.01$. To promote stable convergence, we apply a linear learning rate scheduler with 500 warmup steps. A complete summary of the training hyperparameters is provided in the appendix.

Gradient accumulation is employed to simulate a larger effective batch size, enabling the model to learn from diverse PII contexts without exceeding memory limits. Additionally, gradient clipping with a maximum norm of $1.0$ is applied to prevent exploding gradients during the fine-tuning of the transformer blocks. The training objective is the standard cross-entropy loss computed against the target labels.

\subsection{Fine-tuning Mistral}

To adapt the Mistral-7B model, we employ a parameter-efficient fine-tuning strategy under constrained computational resources. The base Mistral-7B model is loaded using 4-bit quantization with the NF4 scheme via the \texttt{bitsandbytes} library, significantly reducing memory usage while preserving model performance.

The training data consists of instruction-style input-output pairs, where each input contains prompt with unmasked text and the corresponding output represents the same text with personally identifiable information replaced by standardized placeholders. The task is formulated as an instruction-following problem suitable for causal language modeling. Each training instance is constructed using a fixed prompt template, shown below:
\begin{verbatim}
<s>{prompt_text} Input: {input_text} Output: {Output_text}</s>
\end{verbatim}
Due to the large parameter size of Mistral-7B, full fine-tuning is computationally expensive and memory-intensive. To address this, we adopt a parameter-efficient fine-tuning (PEFT) approach using Low-Rank Adaptation (LoRA). LoRA introduces a small number of trainable rank-decomposition matrices into selected attention projection layers, while keeping the original model weights frozen. This approach substantially reduces the number of trainable parameters and enables efficient adaptation of the model without modifying the full parameter space.

The LoRA adapters are integrated using the \texttt{peft} framework and trained using a causal language modeling objective. Training is conducted with mixed-precision arithmetic and gradient accumulation to further improve computational efficiency. The Hugging Face \texttt{Trainer} API is used to manage the training process, including optimization and checkpointing.

After fine-tuning, the learned LoRA adapters are saved independently and later reloaded alongside the frozen base model for inference. The resulting model is evaluated through text generation to verify its ability to correctly identify and mask PII in previously unseen inputs.

\subsection{Implementation Details}
A total of three dataset variants were derived from the AI4Privacy dataset: \textit{normalized}, \textit{replaced}, and \textit{removed}. Each variant contains its own training, validation, and test splits, with 29,718 samples for training and 3,943 samples for both validation and testing. Mistral was finetuned on the \textit{normalized} and \textit{removed} variants with a maximum of 3,000 training steps, while evaluation and testing were performed on the full datasets. The T5-small model was finetuned and evaluated on all three dataset variants. The details about the training parameters are provided in the appendix.

\subsubsection{Training Details}

All experiments were conducted on Linux machines with 2 Tesla T4 GPUs, 2 physical CPUs (4 logical cores), and Python 3.11.13. Mixed-precision (FP16) training was used where applicable.

\begin{table}[h!]
\centering
\caption{Mistral Finetuning Training Details}
\begin{tabular}{lcc}
\toprule
Dataset  & Runtime & GPU(s) \\
\midrule
Normalized  & 11h 55m 50s & Tesla T4 $\times$ 2 \\
Replaced    & 11h 56m 15s & Tesla T4 $\times$ 2 \\
\bottomrule
\end{tabular}
\end{table}

\begin{table}[h!]
\centering
\caption{T5-small Finetuning Training Details}
\begin{tabular}{lcc}
\toprule
Dataset  & Runtime & GPU(s) \\
\midrule
Normalized  & 2h 6m 32s & Tesla T4 $\times$ 2 \\
Replaced    & 2h 13m 8s & Tesla T4 $\times$ 2 \\
Removed     & 2h 6m 16s & Tesla T4 $\times$ 2 \\
\bottomrule
\end{tabular}
\end{table}

\section{Results}
We evaluate our fine-tuned models on a held-out test set of 1,000 examples. We compare our approach against the baselines established in the PRvL benchmark. We analyze performance across three dimensions: span detection capabilities (Span-Correct), strict privacy preservation (Label-Exact), and textual utility (Sequence-Level).

\subsubsection{Span-Detection Evaluation}
First, we evaluate the models' ability to detect PII spans regardless of the entity label. Table \ref{tab:span_correct} compares our models against the PRvL baselines.

Our fine-tuned Mistral 7B (Mistral\_2) achieves good precision (0.963) under relaxed criteria, significantly outperforming the standard Fine-Tuned Llama3.1-8B baseline (0.915) and approaching the state-of-the-art Instruction-Tuned DeepSeek-Q1 (0.973). This indicates that our fine-tuning strategy successfully teaches the model to recognize sensitive spans with high fidelity.

Furthermore, our T5 (Best Checkpoint) implementation vastly outperforms the T5 baseline reported in PRvL. While their T5 model struggled with a precision of 0.727, ours achieves 0.891, demonstrating that encoder-decoder models remain highly effective for span detection when optimized correctly.

\begin{table}[h] \centering \caption{Span-Correct Evaluation. Metrics reflect detection of correct PII spans regardless of label accuracy. Our Mistral model approaches SOTA performance, while our T5 implementation significantly improves upon the PRvL baseline.} 
\label{tab:span_correct} 
\begin{tabular}{lccc} 
\toprule \textbf{Model} & \textbf{Accuracy} & \textbf{Precision} & \textbf{Recall} \\ 
\midrule \multicolumn{4}{l}{\textit{Baselines (PRvL - Span-Correct)}} \\
T5 (Fine-Tuned) & 0.883 & 0.727 & 0.830 \\
Llama3.1-8B (Fine-Tuned) & 0.986 & 0.915 & \textbf{0.969} \\ DeepSeek-Q1 (Inst.) & \textbf{0.994} & \textbf{0.973} & \textbf{0.981} \\ 
\midrule \multicolumn{4}{l}
{\textit{Ours (Relaxed Metrics)}} \\ 
T5 (Best Checkpoint) & 0.972 & 0.891 & 0.910 \\ 
Mistral\_1 & 0.936 & 0.647 & 0.921 \\ 
Mistral\_2 & 0.985 & 0.963 & 0.918 \\ 
\bottomrule 
\end{tabular} 
\end{table}

\subsection{Label-Exact Evaluation}
We first assess the models' ability to correctly identify and classify PII spans. Table presents the results under Strict evaluation criteria, where both the span boundaries and the entity type label must match the ground truth exactly.Our fine-tuned Mistral 7B (Mistral\_2) demonstrates superior performance, achieving an accuracy of 0.985 and a precision of 0.962. This performance surpasses the fine-tuned T5 baseline reported in previous work ($0.884$) and rivals the performance of the instruction-tuned DeepSeek-Q1 model ($0.994$). Notably, the Mistral model exhibits a significantly lower mislabeling rate ($79$) compared to our T5 variants ($413$), suggesting that the decoder-only architecture better retains semantic understanding of entity types (e.g., distinguishing between PERSON and ORG) during the redaction process.While the T5 models (T5\_1, T5\_2, T5\_3) showed high stability across runs, they plateaued at a lower precision ($\approx 0.89$), indicating a tendency to over-redact or misclassify complex spans compared to the Mistral architecture.
\begin{table}[h]
\centering
\caption{Label-Exact Evaluation metrics. \textbf{Mislabel \#} denotes type errors on correctly identified spans. Comparisons are made against PRvL baselines (Fine-Tuned T5 and Instruction-Tuned DeepSeek-Q1).}
\label{tab:label_exact}

\begin{tabular}{lcccc}
\toprule
\textbf{Model} & \textbf{Mislabel \#} & \textbf{Accuracy} & \textbf{Precision} & \textbf{Recall} \\
\midrule
\multicolumn{5}{l}{\textit{Baselines (PRvL)}} \\
T5 (Fine-Tuned) & 1211 & 0.884 & 0.700 & 0.825 \\
DeepSeek-Q1 (Inst.) & 3047 & \textbf{0.994} & 0.945 & \textbf{0.960} \\
\midrule
\multicolumn{5}{l}{\textit{Ours}} \\
T5 (Best Checkpoint) & 413 & 0.971 & 0.889 & 0.908 \\
Mistral\_1 & 78 & 0.936 & 0.645 & 0.920 \\
Mistral\_2 & \textbf{79} & 0.985 & \textbf{0.962} & 0.917 \\
\bottomrule
\end{tabular}
\end{table}

\subsubsection{Sequence Level Evaluation}
Beyond privacy, the redacted text must remain legible and grammatically fluent. We report ROUGE and BLEU scores to measure structural fidelity, and the SPriV score (Specific Privacy Leakage) to quantify the proportion of missed PII tokens ($FN / Total\_Gold\_Tokens$).As shown in Table \ref{tab:sequence_metrics}, Mistral\_2 achieves a BLEU score of 0.911, effectively setting a new state-of-the-art for this dataset, outperforming the previous best of 0.908 (DeepSeek-Q1). This indicates that our fine-tuned Mistral model produces highly natural outputs that preserve the original sentence structure better than encoder-decoder alternatives.\\
We calculated SPriV using the formula $FN / Total\_Gold\_Tokens$. Our fine-tuned Mistral\_2 achieves a leakage rate of 0.0098 ($\approx 1.0\%$), while the best T5 model achieves 0.0115 ($\approx 1.1\%$). Both significantly outperform the standard Fine-Tuned T5 baseline (0.024) from the PRvL paper. Although Mistral\_1 shows a technically lower SPriV (0.0092), this is a result of over-redaction (low precision), making Mistral\_2 the optimal configuration.

\begin{table}[h]
\centering
\caption{Sequence-Level Metrics. \textbf{SPriV} (Specific Privacy Leakage) is lower-is-better. Mistral\_2 achieves the highest BLEU score.}
\label{tab:sequence_metrics}
\begin{tabular}{lccc}
\toprule
\textbf{Model} & \textbf{ROUGE-1 / 2 / L} & \textbf{BLEU} & \textbf{SPriV} $\downarrow$ \\
\midrule
\multicolumn{4}{l}{\textit{Baselines (PRvL)}} \\
T5 (Fine-Tuned) & \textbf{0.940} / 0.857 / \textbf{0.934} & 0.830 & 0.024 \\
DeepSeek-Q1 (Inst.) & 0.915 / 0.846 / 0.915 & 0.908 & \textbf{0.002} \\
\midrule
\multicolumn{4}{l}{\textit{Ours}} \\
T5 (Best Checkpoint) & 0.916 / \textbf{0.905} / 0.915 & 0.810 & 0.011 \\
Mistral\_1) & 0.778 / 0.764 / 0.775 & 0.678 & 0.009 \\
Mistral\_2) & 0.907 / 0.897 / 0.906 & \textbf{0.911} & 0.010 \\
\bottomrule
\end{tabular}
\end{table}

\subsubsection{Stability Analysis}
While Mistral\_2 achieved peak performance, we observed significant variance in stability during fine-tuning. Mistral\_1 serves as a notable failure case. Despite using the same architecture, Mistral\_1 suffered from generation loops and hallucination, evidenced by a Length Ratio of 1.42 (generating 204k tokens against a reference of 143k). This excessive generation caused Precision to drop to 0.645.

In contrast, the T5 models were highly stable across all three runs (T5\_1, T5\_2, T5\_3), with accuracy fluctuating by less than 0.2\%. This suggests that while decoder-only models (like Mistral) offer a higher ceiling for fluency and precision, they are more sensitive to hyperparameter tuning and convergence stability than encoder-decoder architectures in redaction tasks.

\section{Discussion}
Our static evaluation results (Section 4) demonstrate that both T5 and Mistral can achieve near-perfect performance on formal datasets. However, the constraints of real-time PII masking require a balance between accuracy and latency. To evaluate this, we deployed both best-performing models (T5 Best Checkpoint and Mistral Run 2) as backend engines for a Discord bot, processing live messages from a small cohort of real users.

\subsection{Real-World Deployment and Latency}
Table \ref{tab:deployment} illustrates the performance of the models in this live environment. We observed a distinct divergence in practical utility:
\begin{itemize}
\item \textbf{T5 (Efficiency)}: The T5 model proved highly efficient, with an average inference time of 1.46 seconds. This latency is acceptable for near-real-time chat moderation. However, its accuracy dropped to 0.788 on real-world data (down from $\approx$0.97 on the test set). This suggests that the encoder-decoder architecture, while fast, may overfit to the formal structure of the training data and struggle with the noise, slang, and irregular syntax common in Discord conversations.
\item \textbf{Mistral (Robustness)}: The Mistral 7B model demonstrated superior generalization, maintaining a real-world accuracy of 0.876. This indicates that the decoder-only architecture's extensive pre-training on web data allows it to better understand context and identify PII even in informal, noisy text. However, this robustness comes at a severe computational cost: an average inference time of 15.6 seconds. This 10x latency increase makes the 7B model impractical for synchronous chat blocking, though it remains viable for asynchronous tasks like retroactive log sanitization.
\end{itemize}
\begin{table}[h]\centering\caption{Real-World Deployment Metrics (Discord Bot). "Inference Time" is the average latency per message. "Real-World Accuracy" is measured on live user conversations.}
\label{tab:deployment}
\begin{tabular}{lccc}
\toprule\textbf{Model} & \textbf{Inference Time (s)} & \textbf{Real-World Accuracy} \\
\midrule T5 (Best) & \textbf{1.46s} & 0.788  \\
\midrule Mistral (Run 2) & 15.60s & \textbf{0.876}  \\
\bottomrule
\end{tabular}
\end{table}

The results highlight a fundamental trade-off in  PII masking architectures. T5 excels as a lightweight pattern matching architecture offering speed suitable for high-throughput streams but failing when the input deviates from the training distribution. Mistral acts as a better for masking robustness, leveraging its world knowledge to identify PII in ambiguous contexts (e.g., distinguishing a username from a real name in a gaming context) but requires substantial GPU resources.

\section{Conclusion}
In this work, we demonstrated that fine-tuning open-source LLMs can achieve state-of-the-art PII redaction. Our fine-tuned Mistral 7B set a new benchmark, achieving a BLEU score of 0.911 and Label-Exact Precision of 0.962, surpassing existing PRvL baselines.

However, real-world deployment on Discord highlighted a critical trade-off. While Mistral offered superior robustness on noisy text (0.876 accuracy vs. T5's 0.788), its high latency (15.6s) makes it impractical for synchronous use compared to the highly efficient T5 (1.46s). 

Ultimately, high-precision automated redaction is achievable, but the choice of model requires balancing the robust semantic understanding of 7B-parameter models against the low-latency requirements of real-time applications.

\bibliographystyle{unsrtnat}
\bibliography{references} 

@misc{leon2025,
      title={PRvL: Quantifying the Capabilities and Risks of Large Language Models for PII Redaction}, 
      author={Leon Garza and Anantaa Kotal and Aritran Piplai and Lavanya Elluri and Prajit Das and Aman Chadha},
      year={2025},
      eprint={2508.05545},
      archivePrefix={arXiv},
      primaryClass={cs.CR},
      url={https://arxiv.org/abs/2508.05545}, 
}

@article{singh2025,
  title={Unmasking the Reality of PII Masking Models: Performance Gaps and the Call for Accountability },
  author={Devansh Singh and Sundaraparipurnan Narayanan},
  journal={arXiv:2504.12308},
  year={2025}
}

@article{karavdic2025,
  title={HANDLING CONFIDENTIAL DATA IN LLM PROMPTS},
  author={Nedim Karavdić},
  year={2025}
}

@misc{hipaa2025,
  title        = {Summary of the HIPAA Privacy Rule},
  author       = {{U.S. Department of Health and Human Services}},
  year         = {2025},
  month        = mar,
  note         = {Online; accessed 20 June 2025},
  url          = {https://www.hhs.gov/hipaa/for-professionals/privacy/laws-regulations/index.html}
}

@article{ehrmann2021ner_survey,
  title        = {Named Entity Recognition and Classification on Historical Documents: A Survey},
  author       = {Maud Ehrmann and Ahmed Hamdi and Elvys Linhares Pontes and Matteo Romanello and Antoine Doucet},
  journal      = {arXiv preprint arXiv:2109.11406},
  year         = {2021},
  url          = {https://arxiv.org/abs/2109.11406},
  note         = {Also published in *ACM Computing Surveys* 56(2):1–47},
  doi          = {10.48550/arXiv.2109.11406}
}

@article{singh2025_unmasking_pii,
  title        = {Unmasking the Reality of PII Masking Models: Performance Gaps and the Call for Accountability},
  author       = {Devansh Singh and Sundaraparipurnan Narayanan},
  journal      = {arXiv preprint arXiv:2504.12308},
  year         = {2025},
  url          = {https://www.arxiv.org/abs/2504.12308},
  note         = {Accessed 17 December 2025}
}

@article{pearson2021_ner_medical,
  title        = {Named Entity Recognition in Unstructured Medical Text Documents},
  author       = {Cole Pearson and Naeem Seliya and Rushit Dave},
  journal      = {arXiv preprint arXiv:2110.15732},
  year         = {2021},
  url          = {https://arxiv.org/abs/2110.15732},
  doi          = {10.48550/arXiv.2110.15732},
  note         = {Accessed 17 December 2025},
}

@article{rajgarhia2025_hybrid_pii,
  title        = {An Evaluation Study of Hybrid Methods for Multilingual PII Detection},
  author       = {Harshit Rajgarhia and Suryam Gupta and Asif Shaik and Gulipalli Praveen Kumar and Y. Santhoshraj and Sanka Nithya Tanvy Nishitha and Abhishek Mukherji},
  journal      = {arXiv preprint arXiv:2510.07551},
  year         = {2025},
  url          = {https://arxiv.org/abs/2510.07551},
  doi          = {10.48550/arXiv.2510.07551},
  note         = {Accessed 18 December 2025},
}

@inproceedings{vaswani2017_attention,
  title        = {Attention Is All You Need},
  author       = {Ashish Vaswani and Noam Shazeer and Niki Parmar and Jakob Uszkoreit and Llion Jones and Aidan N. Gomez and {\L}ukasz Kaiser and Illia Polosukhin},
  booktitle    = {Advances in Neural Information Processing Systems},
  year         = {2017},
  volume       = {30},
  pages        = {5998--6008},
  organization = {Curran Associates, Inc.},
  url          = {https://proceedings.neurips.cc/paper_files/paper/2017/file/3f5ee243547dee91fbd053c1c4a845aa-Paper.pdf}
}

@article{devlin2018bert,
  title        = {BERT: Pre-training of Deep Bidirectional Transformers for Language Understanding},
  author       = {Jacob Devlin and Ming-Wei Chang and Kenton Lee and Kristina Toutanova},
  journal      = {arXiv preprint arXiv:1810.04805},
  year         = {2018},
  url          = {https://arxiv.org/abs/1810.04805},
  note         = {Accessed 18 December 2025}
}

@misc{he2021debertadecodingenhancedbertdisentangled,
      title={DeBERTa: Decoding-enhanced BERT with Disentangled Attention}, 
      author={Pengcheng He and Xiaodong Liu and Jianfeng Gao and Weizhu Chen},
      year={2021},
      eprint={2006.03654},
      archivePrefix={arXiv},
      primaryClass={cs.CL},
      url={https://arxiv.org/abs/2006.03654}, 
}

@misc{liu2019robertarobustlyoptimizedbert,
      title={RoBERTa: A Robustly Optimized BERT Pretraining Approach}, 
      author={Yinhan Liu and Myle Ott and Naman Goyal and Jingfei Du and Mandar Joshi and Danqi Chen and Omer Levy and Mike Lewis and Luke Zettlemoyer and Veselin Stoyanov},
      year={2019},
      eprint={1907.11692},
      archivePrefix={arXiv},
      primaryClass={cs.CL},
      url={https://arxiv.org/abs/1907.11692}, 
}

@misc{raffel2023exploringlimitstransferlearning,
      title={Exploring the Limits of Transfer Learning with a Unified Text-to-Text Transformer}, 
      author={Colin Raffel and Noam Shazeer and Adam Roberts and Katherine Lee and Sharan Narang and Michael Matena and Yanqi Zhou and Wei Li and Peter J. Liu},
      year={2023},
      eprint={1910.10683},
      archivePrefix={arXiv},
      primaryClass={cs.LG},
      url={https://arxiv.org/abs/1910.10683}, 
}

@misc{jiang2023mistral7b,
      title={Mistral 7B}, 
      author={Albert Q. Jiang and Alexandre Sablayrolles and Arthur Mensch and Chris Bamford and Devendra Singh Chaplot and Diego de las Casas and Florian Bressand and Gianna Lengyel and Guillaume Lample and Lucile Saulnier and Lélio Renard Lavaud and Marie-Anne Lachaux and Pierre Stock and Teven Le Scao and Thibaut Lavril and Thomas Wang and Timothée Lacroix and William El Sayed},
      year={2023},
      eprint={2310.06825},
      archivePrefix={arXiv},
      primaryClass={cs.CL},
      url={https://arxiv.org/abs/2310.06825}, 
}

@misc{touvron2023llamaopenefficientfoundation,
      title={LLaMA: Open and Efficient Foundation Language Models}, 
      author={Hugo Touvron and Thibaut Lavril and Gautier Izacard and Xavier Martinet and Marie-Anne Lachaux and Timothée Lacroix and Baptiste Rozière and Naman Goyal and Eric Hambro and Faisal Azhar and Aurelien Rodriguez and Armand Joulin and Edouard Grave and Guillaume Lample},
      year={2023},
      eprint={2302.13971},
      archivePrefix={arXiv},
      primaryClass={cs.CL},
      url={https://arxiv.org/abs/2302.13971}, 
}

@techreport{nist800122_2010,
  title        = {Guide to Protecting the Confidentiality of Personally Identifiable Information (PII)},
  author       = {Erika McCallister and Tim Grance and Karen Scarfone},
  institution  = {National Institute of Standards and Technology (NIST)},
  type         = {Special Publication 800-122},
  number       = {SP 800-122},
  year         = {2010},
  url          = {https://nvlpubs.nist.gov/nistpubs/legacy/sp/nistspecialpublication800-122.pdf},
  note         = {Guide providing practical guidance on identifying and protecting PII confidentiality},
}

@misc{deshmukh2023lifepiipii,
      title={Life of PII -- A PII Obfuscation Transformer}, 
      author={Ajinkya Deshmukh and Saumya Banthia and Anantha Sharma},
      year={2023},
      eprint={2305.09550},
      archivePrefix={arXiv},
      primaryClass={cs.CL},
      url={https://arxiv.org/abs/2305.09550}, 
}

@inproceedings{tjong2003conll,
  title={Introduction to the CoNLL-2003 shared task: Language-independent named entity recognition},
  author={Tjong Kim Sang, Erik F. and De Meulder, Fien},
  booktitle={Proceedings of CoNLL},
  year={2003}
}

@misc{ai4privacy_pii_masking_200k,
  title        = {{ai4privacy/pii-masking-200k} Dataset},
  author       = {{Ai4Privacy Community}},
  year         = {2023},
  howpublished = {Hugging Face Dataset},
  doi          = {10.57967/hf/1532},
  url          = {https://huggingface.co/datasets/ai4privacy/pii-masking-200k},
  note         = {A synthetic multilingual dataset for training and evaluating PII detection and masking models with annotated spans and text masks, containing ~209,000 examples in multiple languages (English, French, German, Italian).},
}

@inproceedings{carlini2021extracting,
  title     = {Extracting Training Data from Large Language Models},
  author    = {Carlini, Nicholas and Tramer, Florian and Wallace, Eric and Jagielski, Matthew and Herbert-Voss, Ariel and Lee, Katherine and Roberts, Adam and Brown, Tom and Song, Dawn and Erlingsson, {\'U}lfar and Oprea, Alina and Raffel, Colin},
  booktitle = {Proceedings of the 30th USENIX Security Symposium},
  year      = {2021}
}

@inproceedings{brown2020language,
  title     = {Language Models are Few-Shot Learners},
  author    = {Brown, Tom B. and Mann, Benjamin and Ryder, Nick and Subbiah, Melanie and Kaplan, Jared and others},
  booktitle = {Advances in Neural Information Processing Systems (NeurIPS)},
  year      = {2020}
}

@misc{aws_comprehend_2025,
  author       = {{Amazon Web Services}},
  title        = {Natural Language Processing -- Amazon Comprehend},
  year         = {2025},
  howpublished = {Online},
  url          = {https://aws.amazon.com/comprehend/},
  note         = {Accessed: 21 June 2025}
}

@misc{microsoft_presidio,
  author       = {{Microsoft}},
  title        = {Presidio: Data Protection and De-identification SDK},
  howpublished = {GitHub repository},
  url          = {https://github.com/microsoft/presidio},
  note         = {Accessed: 21 June 2025}
}

@misc{google_cloud_dlp_2025,
  author       = {{Google Cloud}},
  title        = {Cloud Data Loss Prevention},
  howpublished = {Online},
  url          = {https://cloud.google.com/security/products/dlp},
  note         = {Accessed: 21 June 2025}
}

@inproceedings{papineni2002bleu,
  title        = {BLEU: a Method for Automatic Evaluation of Machine Translation},
  author       = {Kishore Papineni and Salim Roukos and Todd Ward and Wei-Jing Zhu},
  booktitle    = {Proceedings of the 40th Annual Meeting of the Association for Computational Linguistics},
  pages        = {311--318},
  year         = {2002},
  url          = {https://aclanthology.org/P02-1040/},
}

@misc{lewis2019bartdenoisingsequencetosequencepretraining,
      title={BART: Denoising Sequence-to-Sequence Pre-training for Natural Language Generation, Translation, and Comprehension}, 
      author={Mike Lewis and Yinhan Liu and Naman Goyal and Marjan Ghazvininejad and Abdelrahman Mohamed and Omer Levy and Ves Stoyanov and Luke Zettlemoyer},
      year={2019},
      eprint={1910.13461},
      archivePrefix={arXiv},
      primaryClass={cs.CL},
      url={https://arxiv.org/abs/1910.13461}, 
}

\clearpage
\section*{Appendix}
\appendix

\section{Fine-tuning Hyperparameters}
\begin{table}[h]
\centering
\caption{Fine-tuning Hyperparameters for T5-Small}
\label{tab:t5_hyperparameters}
\begin{tabular}{ll}
\hline\textbf{Parameter} & \textbf{Value} \\ 
\hline\\Base Model Architecture & T5-Small \\
Number of Epochs & 3 \\
Learning Rate & $3 \times 10^{-4}$ \\
Batch Size & 8 \\
Gradient Accumulation Steps & 4 \\
Max Input Sequence Length & 512 tokens \\
Max Target Sequence Length & 512 tokens \\
Optimizer & AdamW \\Weight Decay & 0.01 \\
Warmup Steps & 500 \\
Gradient Clipping Limit & 1.0  \\ 
\hline\end{tabular}
\end{table}

\begin{table}[h]
\centering
\caption{Hyperparameters used for fine-tuning Mistral-7B with LoRA.}
\label{tab:mistral_hyperparameters}
\begin{tabular}{ll}
\toprule
\textbf{Hyperparameter} & \textbf{Value} \\
\midrule
Base model & Mistral-7B-v0.3 \\
Fine-tuning method & LoRA (PEFT) \\
Quantization & 4-bit NF4 \\
Optimizer & AdamW8bit \\
Learning rate & $2.5 \times 10^{-5}$ \\
LoRA rank ($r$) & 32 \\
LoRA scaling factor ($\alpha$) & 64 \\
LoRA dropout & 0.05 \\
Precision & Mixed precision (FP16) \\
Gradient accumulation & Enabled \\
\bottomrule
\end{tabular}
\end{table}

\section{Bot-deployment Examples}

\begin{figure}[H]
    \centering

    \setcounter{subfigure}{0}

    \begin{subfigure}[b]{0.45\textwidth}
        \centering
        \includegraphics[width=\textwidth]{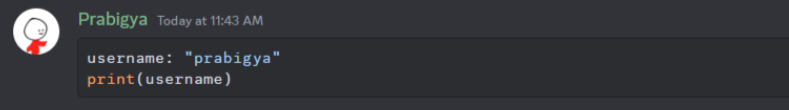}
        \caption{Before}
        \label{fig:ex1a}
    \end{subfigure}
    \hfill
    \begin{subfigure}[b]{0.45\textwidth}
        \centering
        \includegraphics[width=\textwidth]{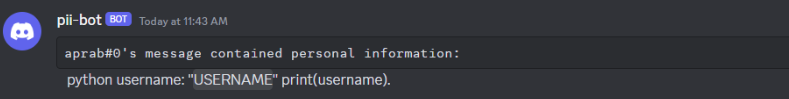}
        \caption{After}
        \label{fig:ex1b}
    \end{subfigure}

    \vspace{0.7em}

    \setcounter{subfigure}{0}

    \begin{subfigure}[b]{0.45\textwidth}
        \centering
        \includegraphics[width=\textwidth]{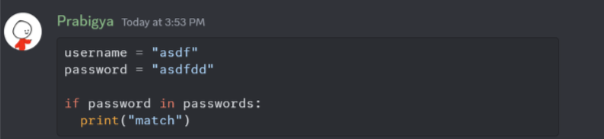}
        \caption{Before}
        \label{fig:ex2a}
    \end{subfigure}
    \hfill
    \begin{subfigure}[b]{0.45\textwidth}
        \centering
        \includegraphics[width=\textwidth]{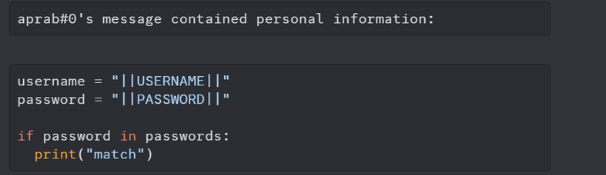}
        \caption{After}
        \label{fig:ex2b}
    \end{subfigure}

    \vspace{0.7em}

    \setcounter{subfigure}{0}

    \begin{subfigure}[b]{0.45\textwidth}
        \centering
        \includegraphics[width=\textwidth]{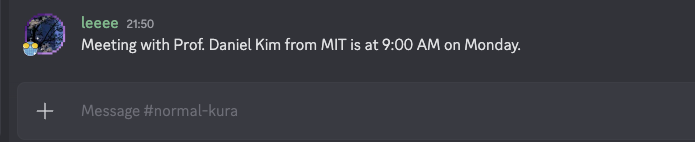}
        \caption{Before}
        \label{fig:ex3a}
    \end{subfigure}
    \hfill
    \begin{subfigure}[b]{0.45\textwidth}
        \centering
        \includegraphics[width=\textwidth]{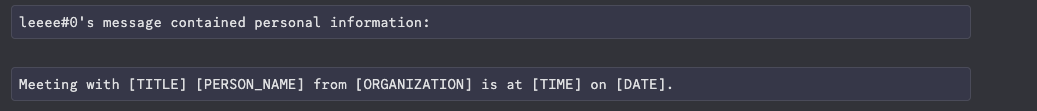}
        \caption{After}
        \label{fig:ex3b}
    \end{subfigure}

    \caption{Discord bot deployment examples demonstrating the model’s masking behavior.
    For each interaction, the original user message (a) and the corresponding masked response (b) are shown.}
    \label{fig:discord_masking}
\end{figure}

\section{Supported PII Labels}
\label{app:pii_labels}

Table~\ref{tab:pii_labels} lists all Personally Identifiable Information (PII)
categories supported by our system and used throughout the experiments.

\begin{table}[h]
\centering
\caption{Supported PII Labels}
\label{tab:pii_labels}
\begin{tabular}{ll}
\toprule
\textbf{Category} & \textbf{PII Label} \\
\midrule
Online Identifiers & \texttt{URL} \\
 & \texttt{USERNAME} \\
 & \texttt{EMAIL} \\
 & \texttt{IP\_ADDRESS} \\
 & \texttt{PASSWORD} \\
\midrule
Personal Attributes & \texttt{PERSON\_NAME} \\
 & \texttt{GENDER} \\
 & \texttt{DATE\_OF\_BIRTH} \\
 & \texttt{TITLE} \\
\midrule
Location Information & \texttt{COUNTRY} \\
 & \texttt{STATE} \\
 & \texttt{CITY} \\
 & \texttt{STREET\_NAME} \\
 & \texttt{BUILDING\_NUMBER} \\
 & \texttt{POSTAL\_CODE} \\
 & \texttt{SECONDARY\_ADDRESS} \\
 & \texttt{GEO\_COORDINATES} \\
\midrule
Temporal Information & \texttt{DATE} \\
 & \texttt{TIME} \\
\midrule
Contact Information & \texttt{PHONE\_NUMBER} \\
\midrule
Organizational Data & \texttt{ORGANIZATION} \\
\midrule
Sensitive Identifiers & \texttt{IDENTIFICATION\_NUMBER} \\
 & \texttt{FINANCIAL\_AMOUNT} \\
 & \texttt{SIGNATURE} \\
\bottomrule
\end{tabular}
\end{table}







\end{document}